\documentclass[11pt]{article}

\usepackage[final]{acl}

\usepackage{times}
\usepackage{latexsym}

\usepackage[T1]{fontenc}

\usepackage[utf8]{inputenc}

\usepackage{microtype}

\usepackage{inconsolata}

\usepackage{graphicx}

\usepackage[utf8]{inputenc} 
\usepackage[T1]{fontenc}    
\usepackage{hyperref}       
\usepackage{url}            
\usepackage{booktabs}       
\usepackage{amsfonts}       
\usepackage{nicefrac}       
\usepackage{microtype}      
\usepackage{xcolor}         

\usepackage{multirow}
\usepackage{comment}
\usepackage{graphicx}
\usepackage{amssymb}
\usepackage{pifont}
\usepackage{makecell}

\usepackage{tabularray}

\usepackage[most]{tcolorbox}
\usepackage{etoolbox}

\usepackage{algorithm}
\usepackage{algpseudocode}
\usepackage{xcolor}

\title{Is GraphRAG Needed?\\ From Basic RAG to Graph-/Agentic Solutions with Context Optimization}

\author{
 \textbf{Long Chen\textsuperscript{1}\thanks{Corresponding authors.}},
 \textbf{Ryan Razkenari\textsuperscript{1}},
 \textbf{Yuxuan Zhou\textsuperscript{1}},
 \textbf{Yuan Tian\textsuperscript{1}},
 \\
 \textbf{Rahul Ghosh\textsuperscript{1}},
 \textbf{Venkatesh Pappakrishnan\textsuperscript{2}},
 \textbf{Disha Ahuja\textsuperscript{2}},
 \textbf{Vidya Sagar Ravipati\textsuperscript{1}\footnotemark[1]}
\\
 \textsuperscript{1}Generative AI Innovation Center, Amazon Web Services (AWS)
 \\
 \textsuperscript{2}Cisco Systems, Inc.
\\
{\tt\small \{longchn, razken, yuxuzh, ytianaws, rahulgh, ravividy\}@amazon.com}
\\
{\tt\small \{vpappakr, disahuja\}@cisco.com}
}
\begin{document}
\maketitle
\begin{abstract}
   As advanced RAG variants like GraphRAG and Agentic RAG emerge, one leading question is when and how to use them. Here, we introduce a framework for different RAG scenarios evaluation and comparison on semi-structured knowledge bases, including regular RAG, GraphRAG, Modular RAG and Agentic RAG. We provide implementation for 9 standardized RAG scenarios, and conduct experiments for a comprehensive comparison. These scenarios are designed for real use cases regarding data and domain restrictions, spanning from simple document-based retrieval to advanced features such as hybrid text-graph retrieval, integration with computed or pre-defined domain knowledge graphs, agentic multi-step planning, and agent-graph integration. Besides, we present a novel context engineering method for GraphRAG and Agentic RAG, addressing the context/memory overflow issues, efficiently managing text and graph retrievals with new representations and agentic loop design, leading to 19\%-53\% reduction on token usage. Moreover, further analysis identifies a retrieval-generation gap where expanded retrieval does not proportionally improve generation quality, suggesting retrieval-oriented metrics overstate advanced retrieval benefits. This work provides data-driven insights on when and how to use them for building production-ready intelligent RAG systems.
   
\end{abstract}

\section{Introduction}

\label{sec:introduction}



Retrieval-Augmented Generation (RAG) \cite{rag_lewis2020retrieval} has become the dominant paradigm for grounding Large Language Model (LLM) outputs in external knowledge. Traditional RAG approaches have proven highly effective for question-answering tasks over unstructured textual corpora. The increasing complexity of real-world applications has highlighted the limitations of purely text-based knowledge representation, driving demand for more sophisticated approaches to handle semi-structured knowledge bases that contain both unstructured textual information and explicit relational data among entities. For instance, precision medicine queries such as "Which gene is involved in vesicle transport, located in the kinetochore, and participates in the antigen processing pathway?" require simultaneous reasoning over entity properties and multi-hop relationships that basic RAG systems struggle to provide.



These needs expose limitations of basic RAG when confronted with semi-structured knowledge bases. Recognizing these limitations, advanced RAG variants have emerged. GraphRAG~\cite{graphrag_peng2024graph} extends traditional retrieval by incorporating graph-based representations and reasoning capabilities, enabling navigation of entity relationships and multi-hop inference over knowledge graphs (KG). Modular RAG~\cite{modularrag_gao2024modular} introduces architectural flexibility by decomposing the pipeline into specialized, interchangeable components optimized for specific data types and query patterns. Agentic RAG systems~\cite{agenticrag_singh2025agentic} leverage autonomous agents capable of dynamic planning, tool utilization, and iterative reasoning, enabling sophisticated multi-step problem-solving that adaptively interacts with various knowledge sources and employs self-correction mechanisms.

However, the rapid adoption of these advanced methodologies raises a fundamental question: do these sophisticated architectures translate to meaningful improvements, and do standard evaluation metrics capture the differences? While these sophisticated systems promise enhanced reasoning capabilities over interconnected data, the practical benefits of these approaches over well-optimized baselines remain largely unexplored.


To address this gap, we introduce a comprehensive framework for evaluating, and comparing different RAG paradigms on semi-structured knowledge bases, encompassing traditional RAG, GraphRAG, Modular RAG, and Agentic RAG approaches. Our work provides systematic implementation of 9 standardized RAG scenarios designed for real-world use cases with realistic data and domain restrictions, spanning from simple document-based retrieval to advanced capabilities including text-to-graph query translation, hybrid text-graph retrieval, integration with computed or pre-defined domain knowledge graphs, agentic multi-step planning, and sophisticated agent-graph integration patterns. Through experimentation on semi-structured knowledge base in the precision medicine domain, we deliver data-driven insights that illuminate when and how to strategically deploy these RAG variants for building production-ready intelligent systems. Besides, we propose a simple but novel context engineering method for GraphRAG and Agentic RAG, addressing the context/memory overflow issues, by leveraging more concise text and graph context representation, as well as a new agentic loop design pattern beyond ReAct~\cite{yao2023react}. We also identify a retrieval-generation gap: end-to-end evaluation of LLM-selected entities during answer generation, rather than raw retrieval rankings, reveals that expanded retrieval does not translate to proportional gains in answer quality. This suggests that retrieval-oriented metrics overstate the benefit of advanced retrieval strategies. Our contributions provide practitioners with empirical guidance for making informed architectural decisions based on specific use case requirements, data characteristics, and performance constraints.

\section{Related works}
\label{sec:related_works}


Semi-structured knowledge bases integrating text and relational data are increasingly important for knowledge-intensive applications. STaRK~\cite{wu2024stark} provides a comprehensive benchmark for evaluating retrieval over such hybrid resources, while datasets like HotpotQA~\cite{yang2018hotpotqa} and ComplexWebQuestions~\cite{talmor2018web} test multi-hop reasoning.


GraphRAG approaches leverage knowledge graphs for relationship-aware retrieval~\cite{edge2024local, graphrag_peng2024graph}, while Agentic RAG introduces autonomous agents that dynamically orchestrate retrieval and reasoning~\cite{agenticrag_singh2025agentic}. The intersection of these approaches with semi-structured knowledge bases presents unique opportunities to leverage both the structural precision of graphs and the flexibility of agent-driven retrieval, though the benefits and trade-offs of these methods compared to simpler baselines remain an open question that our work aims to address.

\section{Methods}
\label{sec:methods}

\subsection{Problem definition}
\label{sec:m_problem}
In this work, we focus on knowledge retrieval and question-answering on semi-structured knowledge bases with both textual and relational information. Following Wu et al.~\cite{wu2024stark}, considering a knowledge base which contains a collection of textual documents $\mathcal{D}$ and a knowledge graph $\mathcal{G = (V, R)}$, where $d_i \in D$ is a textual document describing an entity $ i $, and $v_i \in V$ is the corresponding entity node on KG, with $\mathcal{R}$ being a set of relations between different
nodes. For example, in STaRK-Prime dataset~\cite{wu2024stark}, each entity (e.g., a gene/protein or disease) $ i $ has a textual document $d_i$
that contains the corresponding entity description, and the corresponding node $v_i$ on the knowledge graph $G$ encodes its relations with other nodes such as "side effect", "parent-child", "phenotype present", etc. Given a user query $q$, which could contain certain entity and relation information and requirements, the goal is to retrieve the right set of entities and relations from the knowledge base to answer the user query. 

Thus, this is a typical RAG problem which could be solved by different scenarios. However, the selection of RAG scenarios highly depends on both the availability of the textual and relational resources, as well as the complexity of the user query. In the following sub-sections, we will explain the 9 scenarios under three categories: Regular RAG, GraphRAG, and Modular \& Agentic RAG.

\begin{figure*}[t]
		\centering
		\vspace{-0.5cm}
		\includegraphics[width=0.96\textwidth]{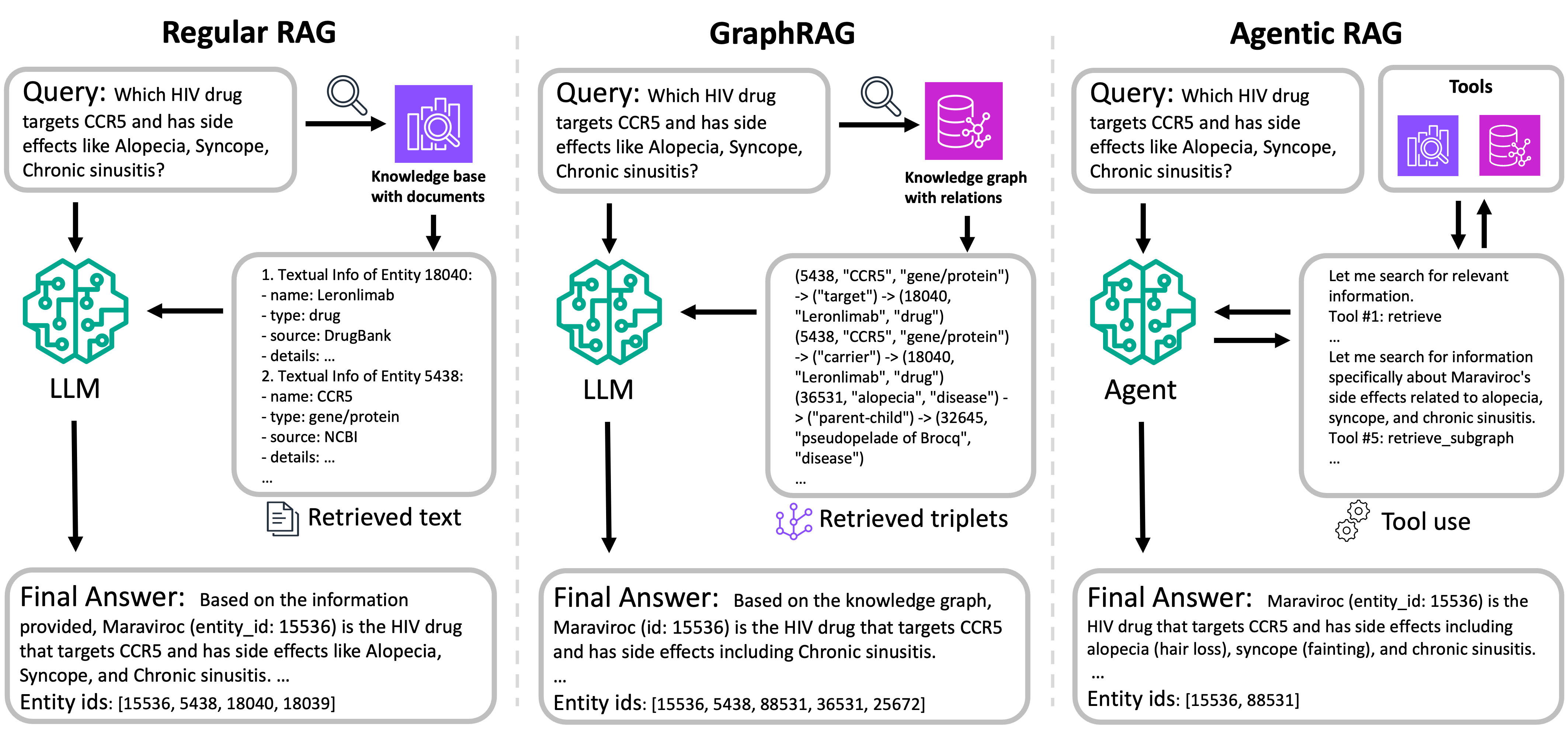}
		\vspace{-0.2cm}\caption{\small {Comparison between Regular RAG, GraphRAG and Agentic RAG.}}\vspace{-0.2cm}
		\label{fig:rag_compare}	
\end{figure*}

\subsection{Regular RAG scenarios}
\label{sec:m_rag}



\textbf{Scenario 1 - RAG with entity description documents only:} This baseline scenario indexes only entity description documents. As shown in Figure \ref{fig:rag_compare}, given a query, text documents are retrieved via vector similarity search and concatenated as context for LLM answer generation. For our task, the LLM is instructed to provide both the answer and the entity ids. This approach represents the most commonly used RAG approach, especially for the case without accessibility to pre-defined relational data. However, it faces limitations such as lack of relational context to handle complex queries.




\textbf{Scenario 2 - RAG with documents containing both entity description and relations:} This scenario augments each entity document with its 1-hop neighbors grouped by relation type before indexing. Figure \ref{fig:doc_example} in Appendix provides examples of the documents in Scenario 1 and 2. The rest of the steps are the same as those in Scenario 1. This provides a simple way to incorporate relational information without graph search, though it is limited to 1-hop relations.

\subsection{GraphRAG scenarios}
\label{sec:m_graph_rag}





\textbf{Scenario 3 - GraphRAG with predefined knowledge graph only:} In this baseline GraphRAG scenario, we utilize solely the predefined knowledge graph without incorporating any textual entity descriptions. The knowledge graph $G = (V, E, R)$ consists of vertices $V$ representing entities, edges $E$ representing relationships, and relation types $R$. The retrieval process operates directly on the graph through the following steps: (1) \textbf{Entity Extraction}: Given a user query $q$, we employ a LLM to identify and extract relevant entity names $E_q = \{e_1, e_2, \ldots, e_m\}$ mentioned in the query. (2) \textbf{Subgraph Extraction}: For each identified entity name $e_i \in E_q$, we perform k-hop traversal from the corresponding graph vertex with a maximum of $N$ paths to extract relevant subgraphs $G_{sub}^{(i)} \subseteq G$. (3) \textbf{Graph Serialization}: The extracted subgraphs are converted into textual representations using triplet format: $\langle subject, predicate, object \rangle$, creating a linearized knowledge context $C_{graph}$. (4) \textbf{Answer Generation}: The serialized subgraph knowledge $C_{graph}$ serves as context for the LLM to generate the final answer as well as the corresponding entity IDs. This scenario establishes the baseline performance when relying exclusively on structured knowledge, without using textual descriptions.


\textbf{Scenario 4 - GraphRAG with computed knowledge graph from entity documents:} Here, we construct a knowledge graph from entity description documents using automated knowledge extraction techniques, combined with vector-based retrieval from the document collection. The methodology consists of the following components: (1) \textbf{Entity Identification}: Apply named entity recognition (NER) techniques to identify entities $E_{doc}$ from the document collection $D = \{d_1, d_2, \ldots, d_n\}$. (2) \textbf{Relation Extraction}: Utilize relation extraction models to identify semantic relationships between co-occurring entities within document contexts, extracting relationship triples $T_{extracted}$. (3) \textbf{Graph Construction}: Build the computed knowledge graph $G_{comp} = (V_{comp}, E_{comp}, R_{comp})$ where vertices and edges are derived from the extracted entities and relationships. (4) \textbf{Graph Refinement}: Apply entity resolution and relation consolidation techniques to merge duplicate entities and standardize relationship types. (5) \textbf{Hybrid Retrieval and Generation}: Combine retrieval results from both the vector index space and the computed knowledge graph for answer generation. This approach evaluates the efficacy of automatically constructed knowledge graphs combined with document-based retrieval.

\textbf{Scenario 5 - GraphRAG with hybrid knowledge integration:} This scenario combines both vector-based document retrieval and predefined knowledge graph traversal. The integration methodology involves: (1) \textbf{Vector-based Entity Retrieval}: Given a user query $q$, perform vector similarity search in the document index space to retrieve candidate entities and their corresponding textual documents $D_{retrieved}$. (2) \textbf{Subgraph Extraction}: For each identified entity name from the retrieved documents, perform h-hop traversal with a maximum of $N$ paths in the predefined knowledge graph to extract relevant subgraphs $G_{sub}$. (3) \textbf{Knowledge Contextualization}: Prepare the extracted subgraphs in triplet format and combine them with the retrieved textual documents to form a comprehensive knowledge context $C_{hybrid} = C_{docs} \cup C_{graph}$. (4) \textbf{Multi-modal Answer Generation}: Provide both the retrieved documents $D_{retrieved}$ and serialized subgraphs $C_{graph}$ to the LLM for answer generation, including the entity IDs. This hybrid approach aims to maximize knowledge coverage by integrating the reliability of curated knowledge graphs with the richness of textual documents.

\subsection{Modular and Agentic RAG Scenarios}



\textbf{Scenario 6 - Modular RAG with well-defined workflow:} This scenario serves as a baseline approach for comparison with subsequent Agentic RAG implementations. It employs a predetermined, sequential workflow to process queries. The modular components are executed in a sequence: (1) \textbf{Query Reformulation}: The input query $q$ is reformulated into an optimized query $q'$ to improve retrieval effectiveness. (2) \textbf{Vector Retrieval}: Perform similarity search in the vector index space of entity documents $D$ to retrieve a set of candidate documents $D_{candidates} = \{d_1, d_2, \ldots, d_k\}$. (3) \textbf{Reranking}: Apply reranking algorithms to the retrieved candidates, producing an ordered list $D_{ranked}$. (4) \textbf{Answer Generation}: Generate the final answer as well as provide the list of selected entities $E_{selected}$ used to answer. This deterministic workflow provides a controlled baseline for evaluating the benefits of increased agent autonomy in subsequent scenarios.


\textbf{Scenario 7 - Agentic RAG with modular components as tools:} In this scenario, all modular components from Scenario 6 are transformed into tools available to an AI agent, which autonomously decides the workflow and tool utilization strategy. The available tool set includes: Query reformatter, Document retriever, and Reranker. The answer generation and entity selection are directly embedded into the agent's system prompt. The agentic workflow operates as follows: (1) \textbf{Autonomous Tool Selection}: Given query $q$, the agent selects appropriate tools from $\mathcal{T}$ based on its reasoning about the query requirements. (2) \textbf{Multi-step Execution}: The agent performs iterative tool calls $\{t_1, t_2, \ldots, t_n\}$ where $t_i \in \mathcal{T}$, with each step informed by previous results. (3) \textbf{Dynamic Query Composition}: Based on intermediate results $R_{i-1}$, the agent can compose new queries $q_i$ for subsequent retrieval steps, enabling adaptive information gathering. (4) \textbf{Convergence Decision}: The agent determines when sufficient information has been gathered to provide a comprehensive answer. This approach enables flexible, context-aware workflows while maintaining access to specialized processing capabilities.

\textbf{Scenario 8 - Autonomous agentic RAG with minimal tools:} This scenario reduces tool availability to test the agent's ability to perform complex reasoning with minimal external support. The agent must internalize processing steps that were previously handled by dedicated tools. The agent configuration includes: (1) \textbf{Single Retrieval Tool}: The agent has access to only one tool $T_{retrieve}$ for vector index search over entity documents. (2) \textbf{Implicit Processing Instructions}: The system prompt instructs the agent to consider processes such as query reformulation, reranking, and entity selection, but provides no specific tools to perform these operations. (3) \textbf{Self-guided Reasoning}: The agent must internally implement logic for: Query analysis and reformulation strategies, Result evaluation and relevance assessment, Entity identification and selection criteria and Information synthesis and answer construction. (4) \textbf{Iterative Refinement}: The agent can perform multiple retrieval iterations, composing new queries $q_i$ based on analysis of previous results $R_{i-1}$, enabling progressive information refinement. This scenario evaluates whether advanced language models can effectively internalize complex retrieval and reasoning processes without explicit tool support.

\textbf{Scenario 9 - Autonomous agentic RAG with knowledge graph retrieval tool:} This scenario extends Scenario 8 by providing the agent with a KG retrieval tool that is the same one as in Scenario 3 and 5. This scenario tests the agent's ability to orchestrate complementary knowledge sources.

\subsection{Context Optimization}
\label{sec:context_opt}
In our experiments with GraphRAG and Agentic RAG systems, we observed significant token usage and cost challenges that frequently approached or exceeded LLM context limits. Analysis revealed two primary sources of redundancy: (1) the conventional triplet format (\texttt{entity1-relation-entity2}) of graph representation repeats entity names when pairs share multiple relations; (2) conventional agentic frameworks like Strands Agents store retrieval results flatly and inject entire session history into each subsequent LLM call, causing overlapping subgraphs and duplicate documents to accumulate in multi-steps.

To address these issues, we propose a three-fold context optimization strategy. First, we introduce a \textbf{relation-grouped graph representation} that transforms conventional triplets into compact format (\texttt{entity1 - (relation1$|$relation2$|$relation3) - entity2}), reducing token usage from $O(n)$ to $O(1)$ for entities connected by $n$ relations. Second, we implement \textbf{graph retrieval deduplication} by maintaining a single unified subgraph throughout the agent session, merging new graph retrievals through entity-level and relation-level deduplication to ensure context grows sublinearly. Third, we implement content-aware \textbf{document retrieval deduplication} using content hashes to identify and eliminate duplicate documents or chunks before adding them to agent memory. 


Beyond deduplication, we further reduce token overhead by modifying the agent's loop design and retrieval pattern. Rather than issuing one query per tool call (the default ReAct~\citep{yao2023react} behavior), we implement \textbf{batch agentic retrieval strategy}, combining ReAct and ReWOO~\citep{xu2025rewoo} style which instructs the agent to formulate multiple complementary sub-queries and execute them in a single batched retrieval call (Algorithm~\ref{alg:batch_agent}). This hybrid ReAct-ReWOO approach reduces the number of LLM round-trips, each of which resends the full conversation history. See Appendix~\ref{app:batch_retrival} for a detailed discussion of the batch strategy's design rationale and its interaction with the retrieval-generation gap.

Moreover, we further investigate the variants of extended retrieval by utilizing the saved token budget such as increasing the max paths and subgraphs, and retrieving nonredundant text chunks. These optimizations also enable controlled experiments to study the retrieval-generation gap, as token savings can be reinvested in expanded retrieval while monitoring whether generation quality follows.  

\begin{algorithm}[th!]
\caption{Batch Agentic Retrieval (Hybrid ReAct-ReWOO Style)}
\label{alg:batch_agent}
\begin{algorithmic}[1]
\Require Query $q$, retrieval tool $\mathcal{T}$, max iterations $K$
\Ensure Answer $a$, entity IDs $E$
\State $\mathcal{M} \gets \emptyset$ \Comment{Agent memory}
\For{$k = 1, \ldots, K$} \Comment{ReAct outer loop}
    \State \textcolor{gray}{\textit{// Think: analyze query and current memory}}
    \State $\{q_1, q_2, \ldots, q_m\} \gets \Call{Plan}{q, \mathcal{M}}$ 
    \State \textcolor{gray}{\textit{// Act: batched retrieval (ReWoo-style)}}
    \State $\mathcal{R}_{\text{batch}} \gets \mathcal{T}(\{q_1, \ldots, q_m\})$ 
        \Comment{Single tool call}
    \State \textcolor{gray}{\textit{// Deduplicate before adding to memory}}
    \State $\mathcal{R}_{\text{new}} \gets \Call{Dedup}{\mathcal{R}_{\text{batch}}, \mathcal{M}}$ 
    \State $\mathcal{M} \gets \mathcal{M} \cup \mathcal{R}_{\text{new}}$
    \State \textcolor{gray}{\textit{// Observe: evaluate sufficiency}}
    \If{$\Call{Sufficient}{q, \mathcal{M}}$}
        \State \textbf{break}
    \EndIf
\EndFor
\State $a, E \gets \Call{Generate}{q, \mathcal{M}}$
\State \Return $a, E$
\end{algorithmic}
\end{algorithm}

\section{Experimental setup}
\label{sec:exp_results}

\textbf{Dataset and metrics:} We evaluate our proposed RAG scenarios on a semi-structured knowledge base dataset: STaRK-Prime~\cite{wu2024stark}. It's a precision medicine inquiry dataset where textual documents are from multiple sources for about 129K entities such as disease, drug, protein and gene, and 8.1M KG relations are from PrimeKG~\cite{primekg_chandak2023building}. We use the official test set of human-generated queries as well as the following evaluation metrics: Hit@1, Hit@5, Recall@20 (R@20), and Mean Reciprocal Rank (MRR). Importantly, unlike the original STaRK benchmark~\cite{wu2024stark}, which evaluates raw retrieval rankings, our metrics assess end-to-end generation quality: the LLM generates a natural language answer and returns a narrowed set of entity IDs, which are then evaluated against ground truth. For context optimization experiments, we include mean Token usage per query (\#Tokens) and Recall of Retrieval (RoR) regarding ground-truth entity IDs coverage before generation utilization.

\textbf{Predefined knowledge graph:} To prepare the predefined KG for graph search, we extracted all the 129K entities and 8.1M relations from STaRK-PRIME dataset, and import them to Amazon Neptune Analytics Graph in openCypher format. Then we used LlamaIndex~\cite{Liu_LlamaIndex_2022} to load the Neptune graph into a property graph index, and used CypherTemplateRetriever to create the knowledge graph retriever. For subgraph extraction, we set h = 2 for h-hop neighbors following Xia et al.~\cite{xia2024knowledge} and maximum of 100 paths to avoid exponentially increasing number of neighbor nodes and relations.

\textbf{Document retrieval:} The document retrievers are built using Amazon Bedrock Knowledge Bases. As the entity description documents have feasible token size, and to keep the completeness of the entity description, no chunking was performed. We used Amazon Titan Text Embedding v2 for embedding, Amazon OpenSearch Serverless as the vector store and information such as the entity type, name as the meta information of the index. For the experiment with the computed knowledge graph from the documents, we used Amazon Neptune Analytics as the vector store and the pre-defined GraphRAG process in Bedrock Knowledge Bases. For each retrieval call, 20 documents are retrieved. 

\textbf{Agentic RAG implementation:} Besides the Document Retrievers and Predefined KG Retriever, we also implemented LLM-based Query Reformatter and Document Reranker for modular RAG (Scenario 6) and corresponding agentic RAG (Scenario 7) experiments. We used Strands Agents~\cite{strands-agents-2025} as the framework for Agentic RAG implementation, with its default ReAct-style~\citep{yao2023react} reasoning loop if not specified, where the agent iteratively thinks about the query, acts by calling a tool, and observes the result before deciding the next step. In Scenario 7, the agent has access to Document Retriever, Query Reformatter and Document Reranker as the tools. While in Scenario 8, the agent only has access to the Document Retriever, but it is instructed in system prompt to consider the other processes autonomously. The system prompt can be found in Appendix Box~\ref{box:prompt_agent_8}. Scenario 9 is similar to Scenario 8, but the agent has access to the Predefined KG Retriever as well.  

\textbf{LLM:} Testing the performance of different LLMs is not the current focus of this work. Thus to fairly compare different RAG scenarios, Anthropic Claude 3.7 Sonnet was used as the core LLM for the LLM-based modules/tools and agents.

\section{Experimental results}
\label{sec:exp_results}

\begin{table*}[!ht]
\caption{Evaluation results for different RAG scenarios.}
  \label{tab:e2e_results}
  \centering

\scriptsize

\begin{tabular}{lcccc}
\toprule
 Methods & Hit@1 & Hit@5 & R@20 & MRR  \\
\midrule
\textbf{Regular RAG} & & & & \\
\textbf{Scenario 1:} RAG with Entity Description Documents only & & & &  \\
- retrieval only (no LLM answer generation and entity selection) & 0.4404 &	0.6697 & 0.7154 & 0.5455 \\
- with LLM answer generation and entity selection & 0.6147	& 0.7523	& 0.6119	& 0.6769 \\

\textbf{Scenario 2:} RAG with Entity Description and Relations Documents & & & &  \\
- retrieval only (no LLM answer generation and entity selection) & 0.5596	& 0.7798	& \textbf{0.7874}	& 0.6647 \\
- with LLM answer generation and entity selection & \textbf{0.6972}	& \textbf{0.8257}	& 0.6728	& \textbf{0.7531} \\

\hline
\textbf{GraphRAG} & & & & \\

\textbf{Scenario 3:} GraphRAG with predefined KG only (max 100 paths) & 0.1376	& 0.1835	& 0.1467	& 0.1542  \\

\textbf{Scenario 4:} GraphRAG with vector search and computed KG from entity documents & & & &  \\
- retrieval only (no LLM answer generation and entity selection) & 0.0000	& 0.1468	& 0.6340	& 0.1036 \\
- with LLM answer generation and entity selection & 0.5596	& 0.6972	& 0.5625	& 0.6162 \\

\textbf{Scenario 5:} GraphRAG with vector search and predefined KG & & & &  \\
- max 100 paths, 5 entity subgraphs & 0.6422 & \textbf{0.7798}	& \textbf{0.6353}	& \textbf{0.7011}	 \\
- max 50 paths, 5 entity subgraphs & 0.6239	& 0.7615	& 0.6061	& 0.6858	 \\
- max 100 paths, 20 entity subgraphs & \textbf{0.6514}	& \textbf{0.7798}	& 0.6131	& \textbf{0.7072} \\
- max 200 paths, 5 entity subgraphs & 0.6147	& 0.7615	& 0.6177	& 0.6831 \\

\hline
\textbf{Modular and Agentic RAG} & & & &  \\
\textbf{Scenario 6:} Modular RAG with well-defined workflow & & & &  \\
- doc retriever only (same as Scenario 1) & 0.6147	& 0.7523	& 0.6119	& 0.6769	 \\
- doc retriever, reranker & 0.5963	& 0.7706	& 0.7140	& 0.6749  \\
- query reformatter, doc retriever, reranker & 0.6239	& 0.8073	& \textbf{0.7857}	& 0.7030 \\
\textbf{Scenario 7:} Agentic RAG with modular components as tools & & & & \\
- tools: query reformatter, doc retriever, reranker & 0.6514 &	0.8165	& 0.7841	& 0.7235 \\
\textbf{Scenario 8:} Autonomous Agentic RAG with minimal tools & & & & \\
- tools: doc retriever & \textbf{0.6881}	& \textbf{0.844}	& 0.7745	& \textbf{0.7549} \\
- tools: doc retriever, enabled thinking  & 0.6296	& 0.8056	& 0.7643	& 0.7160 \\
\textbf{Scenario 9:} Autonomous Agentic RAG with KG retrieval tools & & & & \\
- tools: doc retriever, pre-defined KG retriever & 0.6055 &	0.7890 & 0.7125	& 0.6834 \\

\bottomrule
\end{tabular}
\end{table*}

\begin{table*}
\caption{Evaluation results for methods with context optimization.}
  \resizebox{\linewidth}{!}{
  \label{tab:context_opt_results}
  \centering

\scriptsize

\begin{tabular}{lcccccc}
\toprule
 Methods & Hit@1 & Hit@5 & R@20 & MRR & RoR & \#Tokens  \\
\midrule
\textbf{Scenario 5-Opt:} GraphRAG with vector search and predefined KG + Context opt. & & & & & & \\
- (baseline, no opt) max 100 paths, 5 entity subgraphs & 0.6422 & 0.7798	& \textbf{0.6353}	& 0.7011 &	77.5\% & 49.1K \\

- (with opt) max 100 paths, 5 entity subgraphs & 0.6330 & 0.7890 & 0.6202 & 0.7003 & 77.5\%& \textbf{23.0K(-53\%)}  \\

- (with opt) max 500 paths, 5 entity subgraphs & 0.6422 & \textbf{0.7982} & \textbf{0.6346} & 0.7095 & 79.6\%& 29.0K(-41\%) \\

- (with opt) max 100 paths, 20 entity subgraphs & \textbf{0.6606} & 0.7890 & 0.6074 & \textbf{0.7194} &79.2\% & 33.3K(-32\%)  \\

- (with opt) max 500 paths, 20 entity subgraphs & 0.6055 & 0.7798 & 0.6283 & 0.6835 & \textbf{83.5\%} & 54.3K(+11\%) \\

\hline
\textbf{Scenario 8-Opt:} Autonomous Agentic RAG with minimal tools + Context opt. & & & & & & \\
- (baseline, no opt) tools: doc retriever (20 docs) & \textbf{0.6881}	& \textbf{0.844}	& \textbf{0.7745}	& \textbf{0.7549} & 83.8\% & 237K \\

- (with opt) tools: doc retriever (20 docs)  & 0.6330	& 0.8260	& 0.7470	& 0.7220 & 81.0\%& \textbf{192K(-19\%)} \\

- (with opt) tools: doc retriever (20 nonredundant docs)  & 0.6060	& 0.7800	& 0.7160	& 0.6860 & 86.0\% & 286K(+21\%) \\

- (with opt) tools: doc retriever (20 docs, batch queries)  & 0.6240	& 0.8070	& 0.7610	& 0.7080 & 88.0\% & 197K(-17\%) \\

- (with opt) tools: doc retriever (20 nonredundant docs, batch queries)  & 0.5413	& 0.6972	& 0.6510	& 0.6004 & \textbf{94.4\%} &220K(-7.2\%) \\

\textbf{Scenario 9-Opt:} Autonomous Agentic RAG with KG retrieval tools + Context opt. & & & & & \\
- (baseline, no opt) tools: doc retriever (20 docs), KG retriever (100 paths, 5 subgraphs) & 0.6055 &	0.7890 & 0.7125	& 0.6834 & 84.2\% & 466K \\
- (with opt) tools: doc retriever (20 docs), KG retriever (100 paths, 5 subgraphs) & 0.6240	& 0.8260	& 0.7850	& 0.7120 & 84.7\% & 352K(-24\%) \\
- (with opt) tools: doc retriever (20 docs), KG retriever (500 paths, 5 subgraphs) & \textbf{0.6420}	& \textbf{0.8350}	& \textbf{0.8040}	& \textbf{0.7320} & 84.5\% & 272K(-42\%) \\
- (with opt) tools: doc retriever (20 docs), KG retriever (100 paths, 5 subgraphs), batch queries & 0.5321	& 0.7615	& 0.7603	& 0.6365 & 88.7\% & 264K(-43\%) \\
- (with opt) tools: doc retriever (20 docs), KG retriever (500 paths, 5 subgraphs), batch queries & 0.5229	& 0.7615	& 0.7330	& 0.6296 & \textbf{89.1\%} & \textbf{244K(-48\%)} \\

\bottomrule
\end{tabular}
}
\end{table*}

\begin{table*}[th!]
\caption{Comparison of LLM extracted vs.\ missed answer entities by position metrics.}

\label{tab:position-comparison}
\centering

\scriptsize

\begin{tabular}{lccccccc}
\toprule
\textbf{Group}  & \textbf{In Doc Retrieval} & \textbf{Mean Rank} & \textbf{Med. Rank} & \textbf{Mean Tok \#} & \textbf{Med. Tok \#} & \textbf{Mean Tok \%} & \textbf{Med. Tok \%} \\
\midrule
LLM Extracted & 96.3\% & 5.3 & 3 & 1,888 & 376 & 10.5\% & 2.9\% \\
LLM Missed & 78.2\% & 61.0 & 11 & 5,534 & 5,003 & 36.8\% & 35.9\% \\
\bottomrule
\end{tabular}
\end{table*}

\textbf{Regular RAG:} Table \ref{tab:e2e_results} provides the performance for Regular RAG based Scenarios (Scenarios 1-2). Scenario 1, using only entity description documents with retrieval-only configuration, achieved modest performance with Hit@1 of 0.4404 and MRR of 0.5455. The addition of LLM answer generation and entity selection significantly improved results, boosting Hit@1 to 0.6147 and MRR to 0.6769. Scenario 2, incorporating both entity description and relations documents, showed consistent improvements across all metrics. The retrieval-only approach achieved Hit@1 of 0.5596 and MRR of 0.6647, while the enhanced version with LLM processing reached the highest performance among basic RAG methods with Hit@1 of 0.6972 and MRR of 0.7531. These results clearly demonstrate the effectiveness of incorporating the relation information in a RAG system, even it is conducted in a very simple way as defined in Scenario 2: only adding the 1-hop neighbors and relations to the end of the original documents. Notably, R@20 drops from 0.7154 to 0.6119 when moving from retrieval-only to LLM-selected evaluation, indicating the LLM selects fewer but more precise entities. This divergence between retrieval and generation metrics is a recurring pattern across all scenarios, motivating our generation-aware evaluation approach.

\textbf{GraphRAG:} As shown in the second section of Table \ref{tab:e2e_results}, among 3 GraphRAG scenarios (Scenarios 3-5), Scenario 5 consistently outperformed the others. Scenario 3, using predefined knowledge graphs only, performed poorly across all metrics (Hit@1: 0.1376, MRR: 0.1542), which indicates the challenges of a RAG system purely relying on graph search without utilizing any semantic information. In contrast, the enhanced performance on Scenario 5 over Scenario 3 clearly demonstrated the benefit of utilizing the semantic information in GraphRAG. Scenario 4, which utilized the computed knowledge graph from entity documents, exhibited extreme inconsistency. The retrieval-only configuration failed completely on Hit@1 (0.0000) while achieving competitive R@20 (0.6340). The version with LLM answer generation and entity selection improved Hit@1 to 0.5596 but reduced overall effectiveness (MRR: 0.6162). The performance gap between Scenario 4 and 5 reflected the gap between the computed KG vs predefined KG. In Scenario 4, the computed KG was generated by the extracted entities and relations within the documents, which is challenging to be as accurate and comprehensive as the predefined KG. Moreover, using a low quality computed KG could negatively impact the final RAG performance, which can explain the performance gap between Scenario 4 and Scenario 1. The enhanced performance of Scenario 5 over Scenario 1 proves the benefit of leveraging the relation information from the predefined KG. However, it is interesting to notice that Scenario 2 outperformed Scenario 5, indicating a simple integration of 1-hop relations could lead to better performance than a more sophisticated GraphRAG approach. This result could also due to the non-optimized subgraph representation (triplets) as well as the hyper-parameter selection (e.g., number of hops, maximum number of paths, number of entity for subgraphs). In Scenario 2, the 1-hop relations was grouped by relation types (see example in Figure \ref{fig:doc_example} in Appendix), while in Scenario 5, the subgraphs were represented in triplet format, which often leads to lengthy context with which LLM could lose the attention due to the "lost in the middle" issue~\cite{liu2023lost}. Under Scenario 5, we compared different maximum number of paths and number of entity for subgraphs. It shows the trend of achieving better performance with higher number of entity for subgraphs and maximum number of paths. However, the performance could also drop due to the lengthy context and the "lost in the middle" issue. 

\textbf{Modular and Agentic RAG:} For Modular RAG (Scenario 6), the performance improved while adding new component, which demonstrated the value of component specialization. The complete pipeline with query reformatter, document retriever, and reranker achieved exceptional performance (Hit@1: 0.6239, Hit@5: 0.8073, R@20: 0.7857 MRR: 0.703). Scenario 7 achieved better performance than Scenario 6 with the same modular components (query reformatter, document retriever, and reranker) as tools, which indicates the power of AI agent for self-planning and multi-step reasoning. More interestingly, the autonomous agentic RAG system (Scenarios 8) achieved the best performance under the Modular and Agentic RAG category, which indicates the potential for AI Agent to conduct complex tasks autonomously with minimal human support or predefined specialization tools. However, enabling thinking or adding the predefined KG retriever (Scenario 9) doesn't seem to help. In Appendix \ref{app:case_study}, we shared one example of the thought process of the agent in Scenarios 8, outputted by Strands Agents. That example demonstrates the agent's ability to handle complex, multi-faceted queries requiring extensive domain knowledge integration while maintaining high accuracy. The agent's autonomous reasoning and adaptive behavior enabled it to successfully navigate the complexity of cross-domain medical and environmental relationships. However, the trace also reveals a critical limitation: after several retrieval calls, the accumulated session history repeatedly triggers context window overflow errors, forcing the agent into increasingly narrow searches and ultimately returning incomplete results. This highlights the need for the context optimization methods presented next.

\textbf{Context Optimization:} Table~\ref{tab:context_opt_results}  demonstrates
consistent token reduction across Scenarios 5, 8, and 9 with
context optimization. For Scenario~5 (GraphRAG), optimization
achieves 53\% token reduction with comparable performance at
baseline parameters, while expanding to 500 paths yields the
best Hit@5 (0.7982) with 41\% reduction. For Scenario~8, deduplication alone achieves 19\% reduction with marginal
trade-offs. Most impressively, Scenario 9 demonstrates that optimization benefits scale with retrieval complexity: the baseline configuration achieves 24\% token reduction while improving Hit@5 and R@20, and expanding to 500 paths further enhances performance while achieving token reduction of 42\%. This indicates that context optimization not only reduces overhead but also enhances agents' effective utilization of expanded hybrid text-graph retrievals.

The batch query variant (Algorithm~\ref{alg:batch_agent})
reduces tool calls while increasing retrieval coverage, as shown in both Scenario~8 and 9. It achieves the highest retrieval coverage (RoR: 94.4\%) while still below baseline token usage when combined with nonredundant retrieval. These results demonstrate that batch agentic retrieval, which issues more speculative queries, is an effective approach for increasing retrieval coverage while reducing the number of LLM round-trips and thus the agent memory usage for repeated injection of session history. However, the expanded retrievals do not necessarily improve generation quality as shown in Table~\ref{tab:context_opt_results} across scenarios, directly evidencing the retrieval-generation gap discussed next.

\textbf{Retrieval-Generation Gap:} The performance gains with increasing retrieval scope remain below expectations across scenarios. Our deep-dive analysis (see Appendix \ref{app:context_opt_deep_dive}) reveals that while expanded retrieval significantly increases ground truth entity coverage, LLMs do not effectively leverage this additional information for answer generation, highlighting a critical gap between retrieval comprehensiveness and generation effectiveness.

To understand this gap, we compare LLM-extracted versus missed answer entities in Scenario 5-Opt (500 paths, 20 subgraphs), where retrieval coverage reaches 83.5\% but LLM extraction only 47.9\%. As shown in Table~\ref{tab:position-comparison}, extracted entities appear 3.5× earlier by token position (mean 10.5\% vs. 36.8\%) and the extraction rates drop sharply with position, indicating positional attention decay. Beyond position, qualitative analysis (Appendix~\ref{app:position_analysis}) identifies two additional factors: LLM preference for canonical entities over exhaustive enumeration, and query-induced cardinality expectations suppresses multi-entity extraction.

These findings suggest that current retrieval-oriented metrics (Hit@k, MRR) may overstate the benefit of expanded retrieval, and that evaluation of RAG systems should separately assess retrieval coverage and generation utilization. Besides, the study (Appendix \ref{app:context_opt_deep_dive}) also reveals that the agent with optimized context autonomously calls retrieval tool more, without explicit prompting. These observations merit further investigation.


\section{Conclusions and future work}
\label{sec:conclusion}


We introduce a comprehensive framework for evaluating Regular RAG, GraphRAG, Modular RAG, and Agentic RAG architectures on semi-structured knowledge bases. Our study demonstrates that the question "Is GraphRAG needed?" requires a nuanced, context-dependent answer. While GraphRAG shows promise for complex relational queries, simpler RAG variants can deliver competitive performance with lower overhead, whereas autonomous Agentic RAG with minimal specialized tools achieves superior results across the board. Our context optimization method achieves 19\%-53\% token reduction while maintaining or improving performance, demonstrating practical value for production deployment. Moreover, we identify a retrieval-generation gap that has implications for how RAG systems are evaluated. Our analysis shows that expanded retrieval does not translate to proportional generation improvements, driven by positional attention decay, semantic salience preferences, and query-induced cardinality effects. Future work should address the retrieval-generation gap, through context restructuring and prompt designs encouraging exhaustive extraction. Other directions could develop hybrid approaches balancing efficiency and adaptiveness, explore domain-specific graph utilization strategies.


\section*{Limitations}

Our experiments are conducted on a single dataset (STaRK-Prime) from the precision medicine domain, which may limit generalizability to domains with different structural characteristics or relational complexity. However, the dataset's scale (129K entities, 8.1M relations) and its combination of textual and relational data make it representative of real-world semi-structured knowledge bases, and our framework is designed to be dataset-agnostic. Similarly, all experiments use a single LLM (Anthropic Claude 3.7 Sonnet). While relative scenario performance may shift with different models, our core findings (e.g., the retrieval-generation gap and the effectiveness of simple relational augmentation) reflect architectural properties rather than model-specific behaviors.

The agentic RAG scenarios exhibit inherent non-determinism in tool selection and planning, which can introduce variability despite using temperature zero. Our evaluation is limited to retrieval-oriented metrics (Hit@1, Hit@5, R@20, MRR) due to lack of ground truth natural language answers, and does not assess dimensions such as factual faithfulness or hallucination rates, which we leave for future work. Finally, we did not exhaustively tune hyperparameters across all scenarios, which may leave room for further performance improvements in individual configurations.

\section*{Ethics Statement}


This work uses the publicly available STaRK-Prime dataset~\citep{wu2024stark}, derived from open biomedical knowledge bases including PrimeKG~\cite{primekg_chandak2023building}. No human subjects were involved in our experiments. All evaluations were conducted using automated metrics on existing benchmark queries. Our experiments rely on commercial LLM APIs (Anthropic Claude via Amazon Bedrock), which incur computational costs and associated energy consumption. The biomedical domain application discussed in this work is intended for research evaluation purposes and should not be used for clinical decision-making without proper validation.

\section*{Acknowledgements}
The authors would like to thank Gaurav Rele, Francisco Calderon Rodriguez, and Wan Chen for insightful discussions that helped shape this work. The authors are also grateful to Anila Joshi and Taimur Rashid for their guidance and continued support of this research.


\bibliography{ref}

\clearpage 
\appendix

\section{Agentic RAG Case Study}
\label{app:case_study}

To illustrate the effectiveness of the autonomous agentic RAG approach, we examine a complex query "Why do prostate cancer cases rise in agriculturally intensive areas in California?" from the STaRK-Prime dataset. This query represents a challenging information retrieval task as it requires understanding multiple domains (oncology, agricultural science, and geographical epidemiology) and establishing causal relationships. Here in Box \ref{box:case_study} is the log (thought process) of the agent as outputted by Strands Agents.

\begin{tcolorbox}[title=Box A: Case Study for Agentic RAG, breakable, enhanced]
\label{box:case_study}

\small

\textbf{Query:} Why do prostate cancer cases rise in agriculturally intensive areas in California?

\textbf{Agent:} \\ 
I'll help you find relevant information about the relationship between prostate cancer cases and agriculturally intensive areas in California. Let me search the knowledge base for relevant documents.

\textit{Tool \#1: retrieve} \\ 
Let me search for more specific information about prostate cancer and agricultural areas in California.

\textit{Tool \#2: retrieve} \\ 
Let me search specifically for information about prostate cancer in California agricultural areas.

\textit{Tool \#3: retrieve} \\ 
Let me search specifically for prostate cancer in relation to agricultural areas.

\textit{Tool \#4: retrieve} \\ 
Let me search specifically for information about prostate cancer in California.

\textit{Tool \#5: retrieve} \\ 
Based on my searches, I need to look specifically for prostate cancer information.

\textit{Tool \#6: retrieve} \\ 
\texttt{bedrock threw context window overflow error}

Let me try a more focused search:

\textit{Tool \#7: retrieve} \\ 
\texttt{bedrock threw context window overflow error}

Let me try with a smaller result set:

\textit{Tool \#8: retrieve} \\ 
Let me try a more specific search for prostate cancer:

\textit{Tool \#9: retrieve} \\ 
\texttt{bedrock threw context window overflow error}

Let me try with a more focused search:

\textit{Tool \#10: retrieve} \\ 
\texttt{bedrock threw context window overflow error}

Let me try with a very specific search:

\textit{Tool \#11: retrieve} \\ 
Let me try one more search with a different approach:

\textit{Tool \#12: retrieve} \\ 
\texttt{bedrock threw context window overflow error}

Let me try with a very specific search:

\textit{Tool \#13: retrieve} \\ 
Based on my searches, I've found limited information specifically about prostate cancer in agricultural areas of California. Let me compile what I've found and provide the entity IDs in the required format.

\end{tcolorbox}

As shown in Box \ref{box:case_study}, the agent demonstrated sophisticated reasoning capabilities through multiple stages:

\begin{enumerate}

\item \textbf{Initial Query Decomposition:} The agent first analyzed the query's multiple components: (1) Geographic context (California); (2) Medical condition (prostate cancer); (3) Environmental factor (agricultural intensity); (4) Causal relationship ("why" and "rise")

\item \textbf{Dynamic Search Strategy:} The agent implemented an iterative search process, demonstrating: (1)Progressive refinement of search terms; (2) Context-aware query reformulation; (3)Autonomous decision-making about when to pursue additional information; (4) Self-correction when initial searches yielded insufficient results

\item \textbf{Adaptive Information Integration:} When faced with retrieval challenges, the agent automatically adjusted its approach by: (1) Increasing search specificity; (2) Decomposing complex relationships into manageable components; (3) Synthesizing information across multiple retrievals; (4) Maintaining context coherence across search iterations.
\end{enumerate}

This example demonstrates the system's ability to handle complex, multi-faceted queries requiring extensive domain knowledge integration while maintaining high accuracy. The agent's autonomous reasoning and adaptive behavior enabled it to successfully navigate the complexity of cross-domain medical and environmental relationships. However, it also reveals a critical failure mode: repeated context window overflow errors (occurring in 6 of 13 tool calls), which motivates our context optimization work. This case study underscores the practical necessity of context management for agentic RAG deployment.

\begin{tcolorbox}[title=Box B: Agent System Prompt - Scenario 8, breakable, enhanced]
\label{box:prompt_agent_8}

\small

\texttt{"""} \\
You are an expert information retrieval agent specializing in scientific and medical knowledge.

\texttt{<task>} \\
Your task is to find relevant documents from a knowledge base using the `retrieve' tool. Do NOT try to directly answer queries - focus on finding and returning relevant entity IDs. \\
\texttt{</task>}

\texttt{<instructions>}
\begin{enumerate}
    \item Analyze the query and, if needed, reformulate it to improve search effectiveness.
    \item Then use the retrieve tool to search the knowledge base with the original or reformulated query.
    \item Use the following information when calling the retrieve tool:
    \begin{itemize}
        \item knowledgeBaseId: \{kd\_id\}
        \item numberOfResults: \{num\_results\}
        \item score: 0.0
    \end{itemize}
    \item If you determine that the initial results are insufficient or that follow-up information would be beneficial:
    \begin{itemize}
        \item Formulate additional, more specific queries based on the initial results
        \item Perform follow-up retrievals using these new queries with the same retrieve tool and parameters
        \item Incorporate the new results into your analysis and reranking
    \end{itemize}
    \item Rerank all retrieved results based on their relevance to the original query, combining results from both initial and follow-up retrievals
    \item Return the top UNIQUE entity IDs in the final reranked list. Each ID must come directly from the `entity\_id' field of the retrieved documents.
\end{enumerate}
\texttt{</instructions>}

\texttt{<output\_format>} \\
Your response must be wrapped in \texttt{<json>} tags and follow these requirements:
\begin{itemize}
    \item Only use numeric entity IDs that appear in the `entity\_id' field of retrieved documents
    \item Do not use document names, titles, or other identifiers
    \item Entity IDs must be numbers, not text strings
\end{itemize}

\begin{verbatim}
<json>
{{
"final_answer": "final answer to the user 
question" # The final answer should include 
the 'entity_id' of the entities used to 
answer the user question
"entity_ids_for_final_answer": [1234, 5678] 
# Array of numeric entity IDs ordered by 
relevance
}}
</json>
\end{verbatim}
\texttt{</output\_format>} \\
Return the output in JSON format wrapped in \texttt{<json>} tags as specified above. Do not include any additional text or explanations outside the \texttt{<json>} tags. \\
\texttt{"""}

\end{tcolorbox}

\section{Deep-dive Analysis on Context Optimization Results}
\label{app:context_opt_deep_dive}

Table \ref{tab:context_opt_graph_deep_dive} exhibits the deep-dive analysis on the variants from Scenario 5-Opt (GraphRAG). It reveals a critical disconnect between retrieval comprehensiveness and generation effectiveness in GraphRAG systems. While document retrieval maintains consistent ground truth entity recall at 73.9\% across all variants, expanding KG retrieval parameters dramatically improves entity coverage—from 54.9\% (100 paths, 5 subgraphs) to 83.5\% (500 paths, 20 subgraphs), representing a 52\% relative improvement. However, this substantial increase in retrieved relevant entities does not translate to improved answer quality; LLM answer recall remains nearly flat between 45.4\%-48.2\%, and even slightly decreases with the most comprehensive retrieval configuration (20 subgraphs). This phenomenon suggests that LLMs struggle to effectively identify and synthesize relevant information from expanded graph contexts, potentially due to information overload, difficulty in distinguishing signal from noise in dense graph structures, or limitations in reasoning over large relational contexts. These findings highlight that simply retrieving more graph information is insufficient. Future work could address how to better present, rank, or guide LLMs in utilizing comprehensive graph retrievals for answer generation.

Table~\ref{tab:context_opt_agent_deep_dive} exhibits the deep-dive analysis
for Scenarios~8-Opt and 9-Opt. Token consumption scales
primarily with sequential tool calls rather than retrieval
volume, as each call resends the full conversation history.
Batch strategies reduce tool calls substantially (from 4.2 to
2.8 for standard batch, and to 1.8 for nonredundant batch)
while increasing retrieval coverage (88.0\% and 94.4\%
respectively vs.\ 83.8\% baseline), confirming that retrieval
comprehensiveness does not require many LLM round-trips.

However, batch planning bypasses the intermediate
think-act-observe reasoning of iterative ReAct, causing
broader, more speculative retrieval. Per-query analysis reveals
this directly: the nonredundant batch variant presents 219
documents (128K input tokens) at the final generation step
versus 87 documents (67K tokens) for the iterative nonredundant
variant. Despite achieving 94.4\% retrieval coverage, the
2.5$\times$ larger generation context triggers the
retrieval-generation gap, producing the lowest answer quality
(Hit@1: 0.5413, LLM Answer: 71.1\%). This demonstrates that
batch retrieval's benefit lies in retrieval efficiency (fewer
tool calls, higher coverage per token spent on session
overhead), not in generation quality so far, as the final context size at
the generation step is the critical bottleneck.

This pattern intensifies in Scenario~9, where batch variants
achieve the largest token reductions (43--48\%) but the worst
end-to-end metrics across all Table~\ref{tab:context_opt_results}
configurations (Hit@1: 0.52-0.53), even underperforming the
unoptimized baseline (0.6055). The combined text and graph
retrievals from batch planning create an oversized context that
the LLM cannot effectively process. In contrast, the non-batch
variant (500 paths) achieves 42\% reduction while improving all
metrics over the baseline, confirming that iterative agent
reasoning serves as a critical information filter between
retrieval and generation.

Separating retrieval coverage from LLM utilization across all
agentic variants reveals a consistent pattern: configurations
with higher RoR do not produce higher LLM answer recall. In the test
cases where both iterative and batch configurations retrieved
the same answer entity, ranking position in the final output
differed---with 45 favoring one variant and 58 favoring the
other, despite identical retrieval and temperature$=$0. This
variance in final answer selection, not retrieval, drives
accuracy differences between configurations.

Furthermore, failure analysis identifies three structural
limitations of vector search that neither agentic iteration nor
graph augmentation can address: inverted relationships,
off-label information buried in nested relation fields, and
queries requiring set intersection across structured attributes.
Graph tools provided marginal unique retrieval (+9 answers) but
degraded reranking quality---the additional context diluted
relevance signals, causing answers to rank lower in final output.

\begin{table*}[!ht]
\caption{Deep-dive analysis for different variants of Scenario 5-Opt.}
  \label{tab:context_opt_graph_deep_dive}
  \centering

\scriptsize

\begin{tabular}{lcccc}
\toprule
\multicolumn{1}{c}{}& \multicolumn{4}{c}{Ground Truth Entity Recall} \\
\cmidrule(r){2-5}
 Methods & Doc Retrieval & KG Retrieval &  Doc+KG Retrieval & LLM Answer  \\
\midrule
\textbf{Scenario 5-Opt:} GraphRAG with vector search and predefined KG + Context opt. & & & \\
- (with opt) max 100 paths, 5 entity subgraphs & 73.9\% & 54.9\% & 77.5\% &48.2\% \\

- (with opt) max 200 paths, 5 entity subgraphs & 73.9\% & 58.1\% & 77.8\% &48.2\% \\

- (with opt) max 500 paths, 5 entity subgraphs & 73.9\% & 62.7\% & 79.6\% &47.9\% \\

- (with opt) max 100 paths, 20 entity subgraphs & 73.9\% & 79.2\% & 79.2\% & 45.4\%  \\
- (with opt) max 500 paths, 20 entity subgraphs & 73.9\% & 83.5\% & 83.5\% &47.9\% \\

\bottomrule
\end{tabular}
\end{table*}

\begin{table*}
\caption{Deep-dive analysis for different variants of Scenario 8-Opt and Scenario 9-Opt.}
  \resizebox{\linewidth}{!}{
  \label{tab:context_opt_agent_deep_dive}
  \centering

\scriptsize

\begin{tabular}{lcccc}
\toprule
\multicolumn{1}{c}{}& \multicolumn{1}{c}{}& \multicolumn{1}{c}{}& \multicolumn{2}{c}{Ground Truth Entity Recall} \\
\cmidrule(r){4-5}
 Methods & \#Tool Calls & Latency & Doc+KG Retrieval & LLM Answer  \\
\midrule
\textbf{Scenario 8-Opt:} Autonomous Agentic RAG with minimal tools + Context opt. & & & & \\
- (baseline, no opt) tools: doc retriever (20 docs) & 4.2 & 27.4s & 83.8\% & 77.4\% \\

- (with opt) tools: doc retriever (20 docs)  & 5.3
& 27.6s & 81.0\% & 74.7\%\\

- (with opt) tools: doc retriever (20 nonredundant docs)  & 4.9 & 31.3s & 86.0\% & 71.6\% \\

- (with opt) tools: doc retriever (20 docs, batch queries)  & 2.8 & 22.8s & 88.0\% & 76.1\% \\

- (with opt) tools: doc retriever (20 nonredundant docs, batch queries)  & 1.8 & 19.8s & 94.4\% & 71.1\% \\

\textbf{Scenario 9-Opt:} Autonomous Agentic RAG with KG retrieval tools + Context opt. & & & & \\
- (baseline, no opt) tools: doc retriever (20 docs), KG retriever (100 paths, 5 subgraphs) & 6.7 &71.8s &84.2\%  & 71.2\% \\
- (with opt) tools: doc retriever (20 docs), KG retriever (100 paths, 5 subgraphs) & 8.4 & 43.7s & 84.7\% & 78.6\% \\
- (with opt) tools: doc retriever (20 docs), KG retriever (500 paths, 5 subgraphs) & 7.2	& 35.8s	& 84.5\%	& 76.4\% \\

- (with opt) tools: doc retriever (20 docs), KG retriever (100 paths, 5 subgraphs), batch queries & 2.4 & 39.8s & 88.7\% & 75.7\% \\
- (with opt) tools: doc retriever (20 docs), KG retriever (500 paths, 5 subgraphs), batch queries & 2.3	& 30.4s	& 89.1\%	& 73.6\% \\

\bottomrule
\end{tabular}
}
\end{table*}

\section{Retrieval-Generation Gap: Position Analysis}
\label{app:position_analysis}

To understand why LLMs fail to utilize retrieved entities, we analyze the gap between retrieval coverage (83.5\%) and LLM extraction (47.9\%) in Scenario 5-Opt (500 paths, 20 subgraphs). Of 237 answer entities present in the prompt, the LLM extracts only 136 (57.4\%), missing 101. The prompt structure places retrievals from documents (up to 20 entities with descriptions) first, followed by graph context in triplet format, with an average of 538 unique entities and 14,862 tokens per prompt.

Table~\ref{tab:position-comparison} compares extracted versus missed entities across position metrics. Extracted entities appear 3.5$\times$ earlier by token position (mean 10.5\% vs 36.8\%, or 1,888 vs 5,534 tokens) and are more likely to appear in the Doc retrieval section (96.3\% vs 78.2\%).





Table~\ref{tab:token-decile} shows extraction rates by token position decile. Entities in the first 10\% of tokens achieve an 85.5\% hit rate, dropping sharply to 26.3\% at 30\%--40\% and 0\% at 70\%--80\% and 90\%--100\%. This is consistent with the ``lost in the middle'' phenomenon \citep{liu2023lost}.

\begin{table}[h]
\caption{LLM extraction rate by entity token position decile.}
\label{tab:token-decile}
\centering
\small
\begin{tabular}{lccc}
\toprule
\textbf{Token Position} & \textbf{Hits} & \textbf{Misses} & \textbf{Hit Rate} \\
\midrule
0\%--10\%   & 94 & 16 & \textbf{85.5\%} \\
10\%--20\%  & 14 & 16 & 46.7\% \\
20\%--30\%  & 14 & 12 & 53.8\% \\
30\%--40\%  & 5  & 14 & 26.3\% \\
40\%--50\%  & 3  & 13 & 18.8\% \\
50\%--60\%  & 4  & 10 & 28.6\% \\
60\%--70\%  & 1  & 8  & 11.1\% \\
70\%--80\%  & 0  & 5  & 0.0\% \\
80\%--90\%  & 1  & 4  & 20.0\% \\
90\%--100\% & 0  & 3  & 0.0\% \\
\bottomrule
\end{tabular}
\end{table}

Table~\ref{tab:rank-bins} shows extraction rates by absolute entity rank. Rank-1 entities achieve 97.9\% hit rate, but this drops to 43.5\% for ranks 6--10 and 0\% beyond rank 50. All 136 extracted entities come from the top 50; entities appearing only in the graph triplet section have zero extraction rate.

\begin{table}[h]
\caption{LLM extraction rate by absolute entity rank. The anomalous 62.5\% at ranks 21--50 reflects only 8 samples.}
\label{tab:rank-bins}
\centering
\small
\begin{tabular}{lccc}
\toprule
\textbf{Rank} & \textbf{Hits} & \textbf{Misses} & \textbf{Hit Rate} \\
\midrule
1      & 47 & 1  & \textbf{97.9\%} \\
2      & 20 & 9  & 69.0\% \\
3      & 16 & 4  & 80.0\% \\
4--5   & 16 & 10 & 61.5\% \\
6--10  & 20 & 26 & 43.5\% \\
11--20 & 12 & 29 & 29.3\% \\
21--50 & 5  & 3  & 62.5\% \\
51+    & 0  & 19 & \textbf{0.0\%} \\
\bottomrule
\end{tabular}
\end{table}

\paragraph{Case Study: Gum Disease Query.} To illustrate the interplay between position and semantic relevance, we examine the query: \textit{``I have inflammation in my gums, and it turns swollen and puffy. Which disease could potentially be the issue?''} The ground truth contains 21 disease entities; 18 were retrieved but only 4 were extracted by the LLM. Table~\ref{tab:case-study-gum} shows the retrieved answer entities ordered by rank.

\begin{table}[h]
\caption{Retrieved answer entities for gum disease query, ordered by rank. \checkmark\ = extracted by LLM, \ding{55}\ = missed.}
\label{tab:case-study-gum}

\centering

\scriptsize

\begin{tabular}{clcc}
\toprule
\textbf{Status} & \textbf{Entity (ID)} & \textbf{Rank} & \textbf{Tok \%} \\
\midrule
\checkmark & chronic gingivitis (98905) & 1 & 0.2\% \\
\checkmark & gingivitis (38078) & 2 & 6.0\% \\
\checkmark & necrot.\ ulcer.\ gingivitis (83971) & 3 & 11.6\% \\
\ding{55} & periodontitis, aggressive (27365) & 4 & 17.6\% \\
\ding{55} & periapical periodontitis (36107) & 5 & 23.7\% \\
\ding{55} & suppur.\ periapical period.\ (97566) & 6 & 29.4\% \\
\checkmark & periodontitis (35985) & 7 & 35.0\% \\
\ding{55} & periodontal disease (36476) & 8 & 40.6\% \\
\ding{55} & periodontitis, chronic (28140) & 9 & 41.2\% \\
\ding{55} & stomatitis (36702) & 10 & 46.8\% \\
\ding{55} & apical periodontitis (95458) & 11 & 47.2\% \\
\ding{55} & gingival disease (36118) & 12 & 52.6\% \\
\ding{55} & gingival hypertrophy (95134) & 13 & 52.8\% \\
\ding{55} & pericoronitis (97540) & 15 & 53.8\% \\
\ding{55} & gingival cancer (37557) & 19 & 55.1\% \\
\ding{55} & epulis (39168) & 20 & 68.4\% \\
\ding{55} & acute pericementitis (95367) & 151 & 81.4\% \\
\ding{55} & herpes simplex gingivost.\ (97159) & 229 & 88.7\% \\
\bottomrule
\end{tabular}
\end{table}

The LLM extracted the top 3 entities plus ``periodontitis'' at rank 7 (token position 35.0\%), while missing three subtypes at earlier positions (ranks 4--6, token positions 17.6\%--29.4\%). This reveals that \textbf{semantic relevance acts as a secondary factor}: the LLM prefers canonical, well-known terms (e.g., ``periodontitis'') over specialist subtypes (e.g., ``suppurative periapical periodontitis''), behaving more like a clinician giving a differential diagnosis than exhaustively listing all matches. The ground truth expects exhaustive enumeration, creating a systematic mismatch with LLM generation behavior.

\paragraph{Case Study: Hypertension Drug Query.} We examine a second 
query: \textit{``Imagine you're a pharmacologist on a quest to find an 
experimental drug to treat high blood pressure. What is an 
investigational adrenergic alpha-1 receptor antagonist that has been 
studied for the treatment of hypertension?''} All 14 ground truth 
entities were successfully retrieved into the prompt. However, the 
LLM extracted only 1 entity (Bunazosin, token position 11.1\%), 
despite several valid answers appearing very early in the graph 
section (e.g., Doxazosin at 0.4\%, Guanadrel at 2.8\%, Terazosin at 
4.4\%). The early positions of the missed entities contradict a 
purely position-based explanation. Instead, the singular phrasing 
``What is an investigational...antagonist'' appears to induce a 
\textbf{single-answer cardinality expectation}: the LLM interprets 
the query as requesting one entity rather than an exhaustive list, 
producing only one answer even when many valid ones are present in 
the high-attention zone. This contrasts with the gum disease query's 
plural phrasing (``Which disease...''), where the LLM extracted 
multiple entities. This case reveals a third factor in the 
retrieval-generation gap beyond position and semantic relevance: 
the query's grammatical framing can systematically suppress 
multi-entity extraction.

These findings suggest the retrieval-generation gap stems from three interacting factors: (1) positional attention decay in long contexts, (2) LLM preference for semantically salient entities over exhaustive enumeration, and (3) query-induced cardinality expectations---where singular phrasing (e.g., ``What is...'') causes the LLM to output a single entity even when multiple valid answers exist in the context. Simply expanding retrieval scope is insufficient; future work should investigate how to structure and present retrieved information to maximize LLM utilization.

\begin{figure*}[tp]
		\centering
		\includegraphics[width=0.96\textwidth]{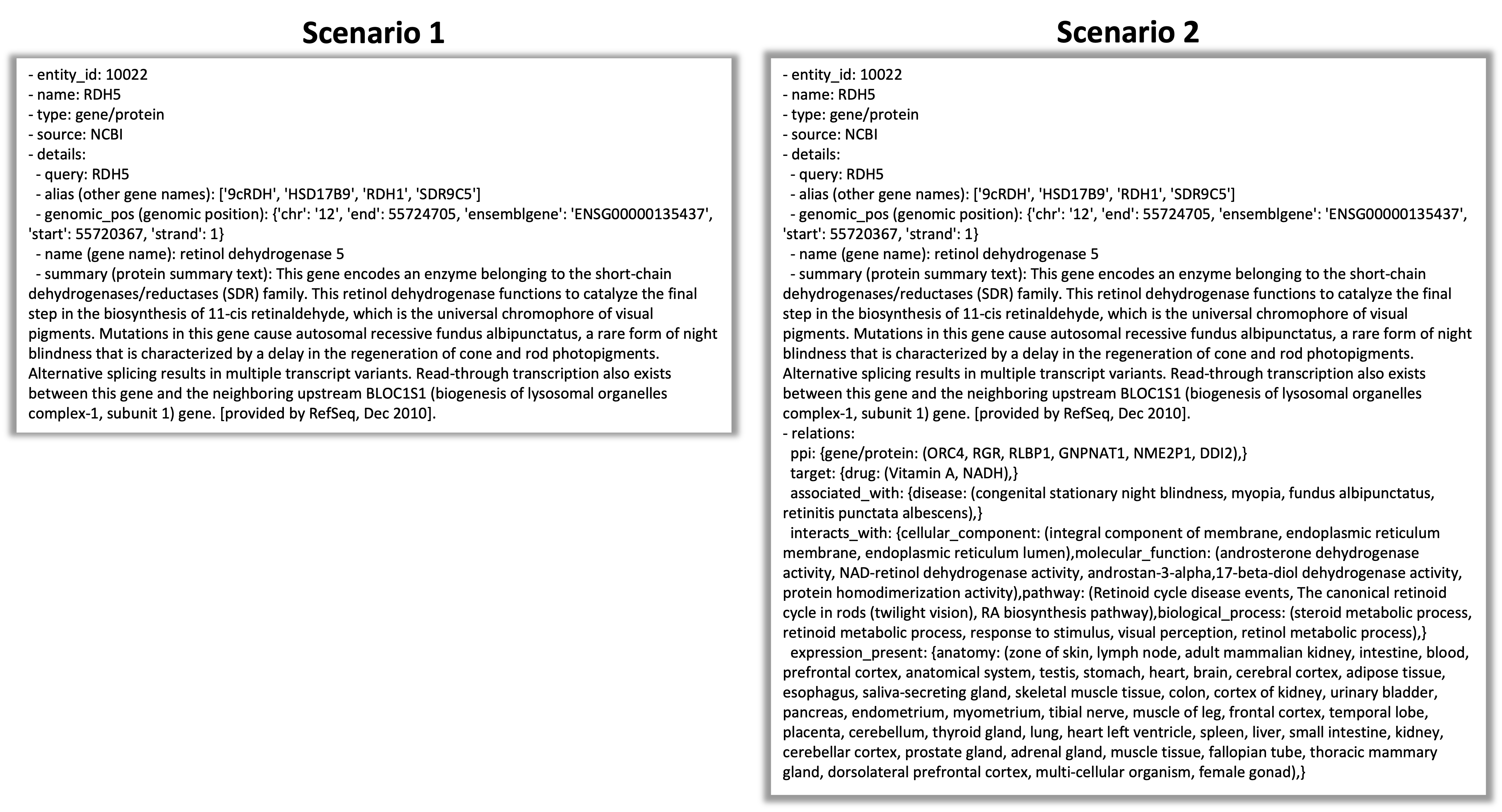}
        \caption{\small {Example of the documents used in Scenario 1 and 2.}}
		\label{fig:doc_example}	
\end{figure*}

\section{Batch Agentic Retrieval Strategy}
\label{app:batch_retrival}
In our analysis of agentic RAG token consumption, we identified that the dominant cost driver is sequential tool calls rather than retrieval volume: each call resends the full conversation history to the LLM, so a 5-call session consumes approximately 250K tokens while a 3-call session uses only 36K regardless of documents retrieved per call. This observation motivates a hybrid retrieval strategy that combines elements of 
ReAct~\citep{yao2023react} and ReWOO~\citep{xu2025rewoo}.

The default agent implementation in Strands Agents follows a standard ReAct-style loop: the agent \textit{thinks} about the query, \textit{acts} by calling a single tool with a single query, \textit{observes} the result, and repeats. In contrast, our batch variant (Algorithm~\ref{alg:batch_agent}) modifies the inner action step: 
instead of issuing one retrieval query per tool call, the agent formulates multiple sub-queries simultaneously and executes them in a single batched tool call. The outer loop remains ReAct-style (iterative think-act-observe with a sufficiency check), while the inner action step follows the ReWOO principle of planning multiple retrieval actions before execution.

Concretely, at each iteration the agent is instructed to decompose the information need into multiple complementary sub-queries (e.g., targeting different entities, 
relations, or constraints mentioned in the original query). These sub-queries are passed to the retrieval tool in a single call, and the results are deduplicated using the content-aware hashing described in Section~\ref{sec:context_opt} before 
being added to the agent's memory. This reduces the number of LLM round-trips from an average of 4.2 sequential calls to 2.8 batched calls, while \textit{increasing} retrieval coverage from 84\% to 88\% (Table~\ref{tab:context_opt_agent_deep_dive}), demonstrating that retrieval comprehensiveness does not require many LLM round-trips.

This design reduces the number of LLM round-trips, each of which resends the full conversation history (Appendix~\ref{app:context_opt_deep_dive}). However, batch planning bypasses the intermediate reasoning of iterative ReAct: rather than evaluating and narrowing queries after each retrieval, all sub-queries are issued upfront based on initial analysis alone. This produces
broader, more speculative retrieval that increases coverage but inflates the final generation context. When combined with nonredundant deduplication or multi-source retrieval (Scenario~9), the accumulated context exceeds the LLM's effective attention span, widening the retrieval-generation gap despite higher
retrieval coverage (see Appendix~\ref{app:context_opt_deep_dive} for detailed analysis).


\end{document}